\documentclass[10pt,twocolumn,letterpaper]{article}

\usepackage{cvpr}
\usepackage{times}
\usepackage{epsfig}
\usepackage{amsmath}
\usepackage{amssymb}
\usepackage{xcolor}
\usepackage{graphicx}
\usepackage{booktabs}
\usepackage{breqn}
\usepackage{caption}
\usepackage{dblfloatfix}
\usepackage{array}
\usepackage{balance}
\usepackage{adjustbox}
\usepackage{multirow}
\usepackage{amsbsy}
\usepackage{subfig}
\usepackage{float}
\usepackage{enumitem}

\usepackage[pagebackref=true,breaklinks=true,letterpaper=true,colorlinks,bookmarks=false]{hyperref}

\cvprfinalcopy 

\def\cvprPaperID{1903} 

\newcommand{\PAR}[1]{\vskip3pt \noindent{\bf #1~}}

\ifcvprfinal\fi
\begin{document}

\title{\vspace{-40pt}Neural Sequential Phrase Grounding (SeqGROUND)}

\author{Pelin Dogan\textsuperscript{1,3} \quad Leonid Sigal\textsuperscript{2} \quad Markus Gross\textsuperscript{1,3}\\
	\textsuperscript{1}ETH Z{\"u}rich \quad \textsuperscript{2}University of British Columbia 
	\quad \textsuperscript{3}Disney Research
	\\%
	{\tt\small{\{pelin.dogan, grossm\}@inf.ethz.ch}, lsigal@cs.ubc.ca}}

\maketitle

\begin{abstract}
We propose an  end-to-end approach for phrase grounding in images. Unlike prior methods that typically attempt to ground each phrase independently by building an image-text embedding, our architecture formulates grounding of multiple phrases as a sequential and contextual process. Specifically,  we encode region proposals and all phrases into two stacks of LSTM cells, along with so-far grounded phrase-region pairs. These LSTM stacks collectively capture context for grounding of the next phrase. The resulting architecture, which we call SeqGROUND, supports many-to-many matching by allowing an image region to be matched to multiple phrases and vice versa. We show competitive performance on the Flickr30K benchmark dataset and, through ablation studies, validate the efficacy of sequential grounding as well as individual design choices in our model architecture.
\end{abstract}

\section{Introduction}

\begin{figure}
	\centering
	\includegraphics[width=0.85\columnwidth]{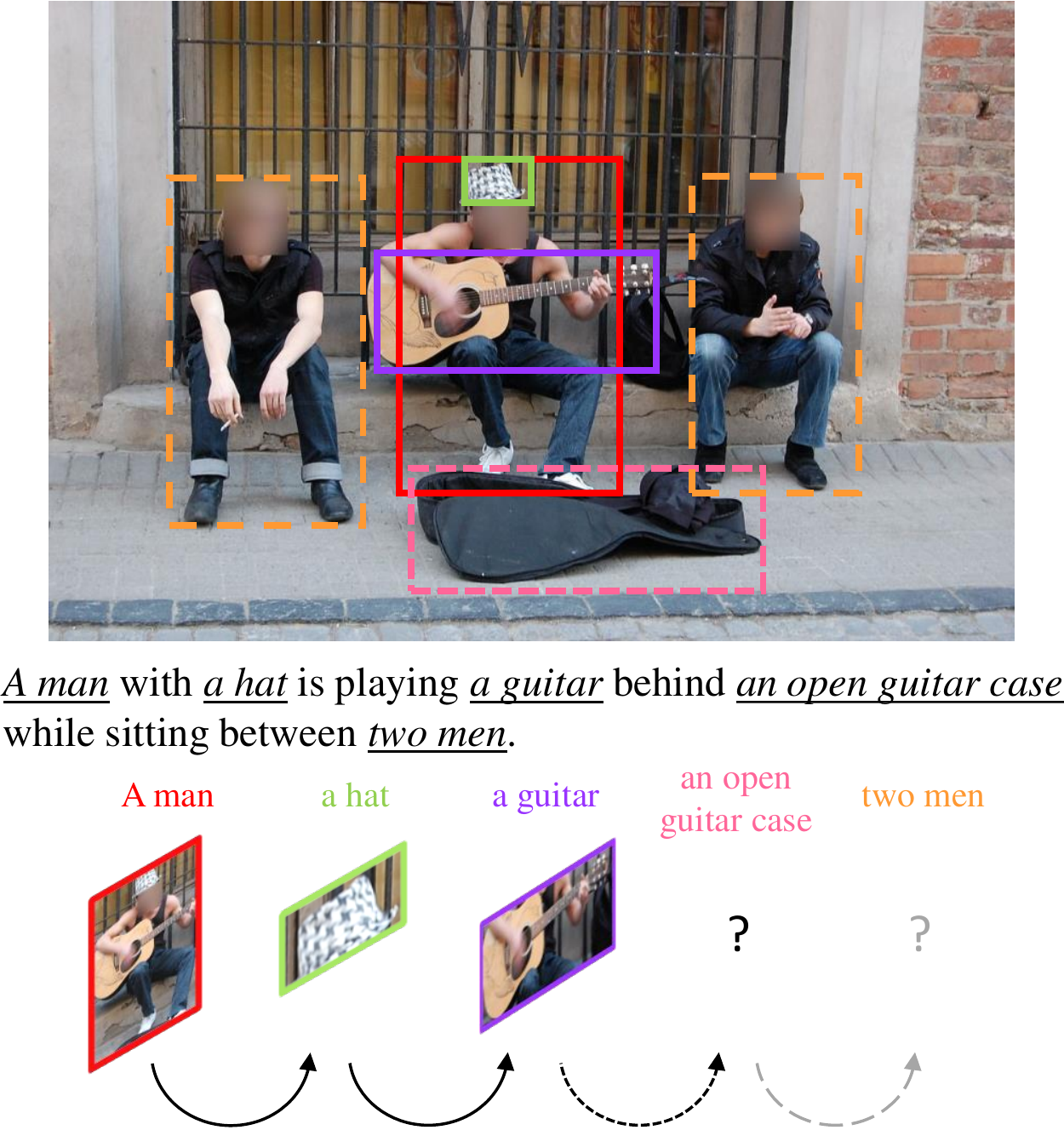}
	\caption{{\bf Illustration of SeqGROUND.} The proposed neural architecture performs phrase grounding sequentially. It uses the previously grounded phrase-image content to inform the next grounding decision (in reverse {\em lexical} order).}
	\label{fig:teaser}
	\vspace{-9pt}
\end{figure}

In recent years, computer vision has made significant progress in standard recognition tasks, such as 
image classification \cite{krizhevsky2012imagenet}, 
object detection \cite{Redmon2016CVPR, ren2015faster}, 
and segmentation \cite{chen2016cvpr}; as well as in more expressive 
tasks that combine language and vision. 
Phrase grounding \cite{plummer2017phrase,wang2016structured,xiao2017cvpr,zhang2017cvpr}, 
a task of localizing a given natural language phrase in an image, has recently gained research attention. 
This constituent task, that generalizes object detection/segmentation, has a breadth of applications that span 
image captioning \cite{johnson2016cvpr,karpathy2015cvpr,xu2015show}, 
image retrieval \cite{gordo2016eccv}, 
visual question answering \cite{antol2015vqa,fukui2016emnlp,tommasi2016bmvc}, 
and referential expression generation \cite{hu2016natural,kazemzadeh2014emnlp,lue2017cvpr,mao2016cvpr}. 

While significant progress has been made in phrase grounding, stemming from release of several 
benchmark datasets \cite{kazemzadeh2014emnlp,krishna2017ijcv,mao2016cvpr,plummer2015flickr30k} 
and various neural algorithmic designs, the problem is far from being solved. 
Most, if not all, existing phrase grounding models can be categorized into two classes: 
attention-based \cite{xiao2017cvpr} or region-embedding-based \cite{plummer2018conditional,zhang2017cvpr}. 
In the former, neural attention mechanisms are used to localize the phrases by, typically, predicting a 
course-resolution mask (\eg, over the last convolutional layer of VGG \cite{simonyan2014very} or another 
CNN network \cite{he2016deep}). In the latter, the traditional object detection 
paradigm is followed by first detecting proposal regions and then measuring a (typically learned) similarity 
of each of these regions to the given language phrase. Importantly, both of these classes of models 
consider grounding of individual phrases individually (or independently), lacking the ability to take into 
account visual and, often, lingual context and/or reasoning that may exist among multiple constituent phrases.

Consider image grounding noun phrases from a given sentence: ``{\em \underline{A lady} sitting on 
\underline{a colorful decoration} with \underline{a bouquet of flowers}, that match \underline{her hair}, 
in \underline{her hand}.}" Note that while multiple {\em ladies} may be present in the image, the grounding 
of ``{\em \underline{a colorful decoration}}" uniquely disambiguates to which of these instances the phrase 
``{\em \underline{A lady}}" should be grounded to. While contextual reference in the above example is spatial, 
other context, including visual maybe useful, \eg, between ``{\em \underline{her hair}}" and 
``{\em \underline{a bouquet of flowers}}".

Conceptually similar contextual relations exist in object detection and have just started to be explored 
through the use of spatial memory \cite{chen2017iccv} and convolutional graph networks (CGNNs) 
\cite{chen2018cvpr,yang2018eccv}. Most assume 
orderless graph relationships among objects with transitive reasoning. In phrase grounding, on the other 
hand, the sentence, from which phrases are extracted, may provide implicate linguistic space- and 
time-order \cite{hazenBook}. We show that such ordering is useful as a proxy for sequentially contextualizing 
phrase grounding decisions. In other words, the phrase that appears {\em last} in the sentence is grounded 
first and is used as context for the next phrase grounding in {\em reverse} lexical order. 
This explicitly sequential process is illustrated in Figure~\ref{fig:teaser}. To our knowledge, our paper is 
the first to explore such sequential mechanism and architecture for phrase grounding. 

Expanding on the class of recent temporal alignment networks (\eg, NeuMATCH \cite{dogan2018neural}), 
that propose neural architectures where discrete alignment actions are implemented by moving data 
between stacks of Long Short-term Memory (LSTM) blocks, we develop a sequential {\em spatial} phrase 
grounding network that we call SeqGROUND. SeqGROUND encodes region proposals and all phrases 
into two stacks of LSTM cells, along with so-far grounded phrase-region pairings. These LSTM stacks 
collectively capture the context for the grounding of the next phrase.  

\vspace{0.1in}
\noindent
{\bf Contributions.} The contributions of this paper are three-fold. First, we propose the notion of contextual 
phrase grounding, where earlier grounding decisions can inform the latter.  Second, we formalize this 
process in the end-to-end learnable neural architecture we call SeqGROUND. The benefit of this 
architecture is its ability to sequentially process many-to-many grounding decisions and utilize rich context of prior 
matches along the way. Third, we show competitive performance both with respect to the prior 
state-of-the-art and ablation variants of our model. Through ablations we validate the efficacy of 
sequential grounding as well as individual design choices in our model. 

\section{Related Work}
Localizing phrases in images by performing sequential grounding is related to multiple topics in multi-modal learning. We briefly review the most relevant literature. 

\PAR{Multi-modal Text and Image Tasks.}
Popular research topics in multi-modal learning include image captioning \cite{karpathy2015deep, mao2014deep, vinyals2015show, xu2015show}, retrieval of visual content~\cite{lin2014visual}, text grounding in images \cite{fukui2016multimodal, plummer2017phrase, rohrbach2016grounding, wang2018learning} and visual question answering \cite{antol2015vqa, sadeghi2015viske, xu2016ask}.
Most approaches along these lines can be classified as belonging to either (i) joint language-visual embeddings or (ii) encoder-decoder architectures. 

The joint {\em vision-language embeddings} facilitate image/video or caption/sentence retrieval by learning to embed images/videos and sentences into the same space \cite{pan2016jointly, torabi2016learning, Xu2017Emotion, xu2015jointly}. For example, \cite{hodosh2013framing} uses simple kernel CCA and in \cite{farhadi2010every} both images and sentences are mapped into a common semantic {\em meaning} space defined by object-action-scene triplets. More recent methods directly minimize a pairwise ranking function between positive image-caption pairs and contrastive (non-descriptive) negative pairs; various ranking objective functions have been proposed including max-margin \cite{kiros2014unifying} and order-preserving losses \cite{vendrov2015order}. 
The {\em encoder-decoder} architectures \cite{torabi2016learning} are similar, but instead attempt to encode images into the embedding space from which a sentence can be decoded. 

Of particular relevance is NeuMATCH \cite{dogan2018neural}, an architecture for video-sentence alignment, where discrete alignment actions are implemented by moving data 
between stacks of Long Short-term Memory (LSTM) blocks. 
We generalize the formulation in \cite{dogan2018neural} to address the spatial grounding of phrases. This requires addition of the spatial proposal mechanism, 
modifications to the overall architecture in order to allow many-to-many matching, modification to the loss function and a more sophisticated training procedure. 

\PAR{Phrase Grounding.}
Phrase grounding, a problem addressed in this paper, is defined as spatial localization of the natural language phrase in an image. A number of approaches have been proposed for grounding over the years. 
 
Karpathy \emph{et al.} \cite{karpathy2014deep} propose to align sentence fragments and image regions in a subspace. 
Rohrbach \emph{et al.}~\cite{rohrbach2016grounding} propose a method to learn grounding in images by reconstructing a given phrase using an attention mechanism. 
Fukui \emph{et al.}~\cite{fukui2016multimodal} uses multimodal compact bilinear pooling to represent multimodal features jointly which is then used to predict the best candidate bounding box in a similar way to~\cite{rohrbach2016grounding}.
Wang \emph{et al.}~\cite{wang2016learning} learns a joint image-text embedding space using a symmetric distance function which is then used to score the bounding boxes to predict the closest to the given phrase. In~\cite{wang2018learning}, their embedding network is extended by introducing a similarity network which aggregates multimodal features into a single vector rather than an explicit embedding space.   
Hu \emph{et al.}~\cite{hu2016natural} proposes a recurrent neural network model to score the candidate boxes using local image descriptors, spatial configurations, and global scene-level context. 
Plummer \emph{et al.}~\cite{plummer2017phrase} perform global inference using a wide range of image-text constraints derived from attributes, verbs, prepositions, and pronouns.
Yeh \emph{et al.}~\cite{yeh2017interpretable} uses word priors with the combination of segmentation masks, geometric features, and detection scores to select the candidate bounding box. 
Wang \emph{et al.}~\cite{wang2016structured} proposes a structured matching method which attempts to reflect the semantic relation of phrases onto the visual relations of their corresponding regions without considering the global sentence-level context.
Plummer \emph{et al.}~\cite{plummer2018conditional} proposes to use multiple text-conditioned embeddings in a single end-to-end model with impressive results on 
Flickr30K Entities dataset~\cite{plummer2015flickr30k}.

These existing works ground each phrase independently, ignoring the semantic and spatial relations among the phrases and corresponding regions respectively. A notable
exception is the approach of Chen  \emph{et al.}~\cite{chen2017query}, where a query-guided regression network, designed to regress the rank of candidates phrase-region pairings,
is proposed along with a reinforcement learning context policy network for contextual refinement of this ranking. 
For \textit{referring expression comprehension}, which is closely related to \textit{phrase grounding} problem, \cite{yu2016modeling, nagaraja2016modeling, yu2018mattnet} introduce taking account of context. Regarding visual data, they consider local context provided by the surrounding objects only. In addition, \cite{nagaraja2016modeling, yu2018mattnet} use textual context with an explicit structure, based on the assumption that “referring expressions mention an object in relation with some other object”. On the other hand, our method represents visual and textual context in a less structured, but more global, manner which alleviates more explicit assumptions made by other methods. Importantly, unlike \cite{yu2016modeling, nagaraja2016modeling, yu2018mattnet}, it makes use of prior matches through a sequential decision process.
In summary, existing approaches perform phrase grounding with two constraints: a region should be matched to no more than one phrase, or a phrase should be matched to no 
more than one region. Furthermore, most of these approaches consider the local similarities rather taking account both global image-level and sentence-level context. 
Here we propose an end-to-end differentiable neural architecture that considers all possible sets of bounding boxes to match any phrase in the caption, and vice versa. 

\section{Approach}
\label{sec:approach}
We now present our neural architecture for grounding phrases in images.
We assume that we need to ground multiple, potentially inter-related, phrases in each image.
This is the case for the Flickr30k Entities dataset, where phrases/entities come from sentence parsing.
Specifically, we parse the input sentence into a sequence of phrases $\mathcal{P} = \left\{P_j\right\}_{j=1\ldots N}$ 
keeping the sentence order; \ie $j=1$ is the first phrase and $j=N$ is the last. 
For a typical sentence in Flickr30k, $N$ is between $1$ and $54$. 
The input image $I$ is used to extract region proposals in the form of bounding boxes.
These bounding boxes are ordered to form a sequence 
$\mathcal{B} = \left\{B_i\right\}_{i=1\ldots M}$. We discuss the ordering choices, 
for both $\mathcal{P}$ and $\mathcal{B}$, and their effects in Section~\ref{subsec:ablations}. 
Our overall task is to ground phrases in the image by matching them to their corresponding 
bounding boxes, for example, finding a function $\pi$ that maps an index of the phrase to its 
corresponding bounding boxes $\langle P_j, B_{\pi(j)} \rangle$. Our method allows many-to-many 
matching of the aformentioned input sequences. 
In other words, a single phrase can be grounded to multiple bounding boxes, or multiple phrases 
of the sentence can be grounded to the same bounding box. 

Phrase grounding is a very challenging problem exhibiting the following characteristics. 
First, image and text are heterogeneous surface forms concealing the true similarity structure.
Hence, satisfactory understanding of the entire language and visual content is needed for effective grounding. 
Second, relationships between phrases and boxes are complex. It is possible (and likely) to have 
many-to-many matchings and/or unmatched content (due to either lack of precision in
the bounding box proposal mechanism or hypothetical linguistic references).
Such scenarios need to be accommodated by the grounding algorithm. 
Third, contextual information that is needed for learning the similarity between phrase-box pairs 
are scattered over the entire image and the sentence. 
Therefore, it is important to consider all visual and textual context with a strong representation of their 
dependencies when making grounding decisions, and to create an end-to-end network, where gradient from grounding decisions can inform content understanding and similarity learning.
 
\begin{figure*}[t]
	\centering
	\includegraphics[width=2\columnwidth]{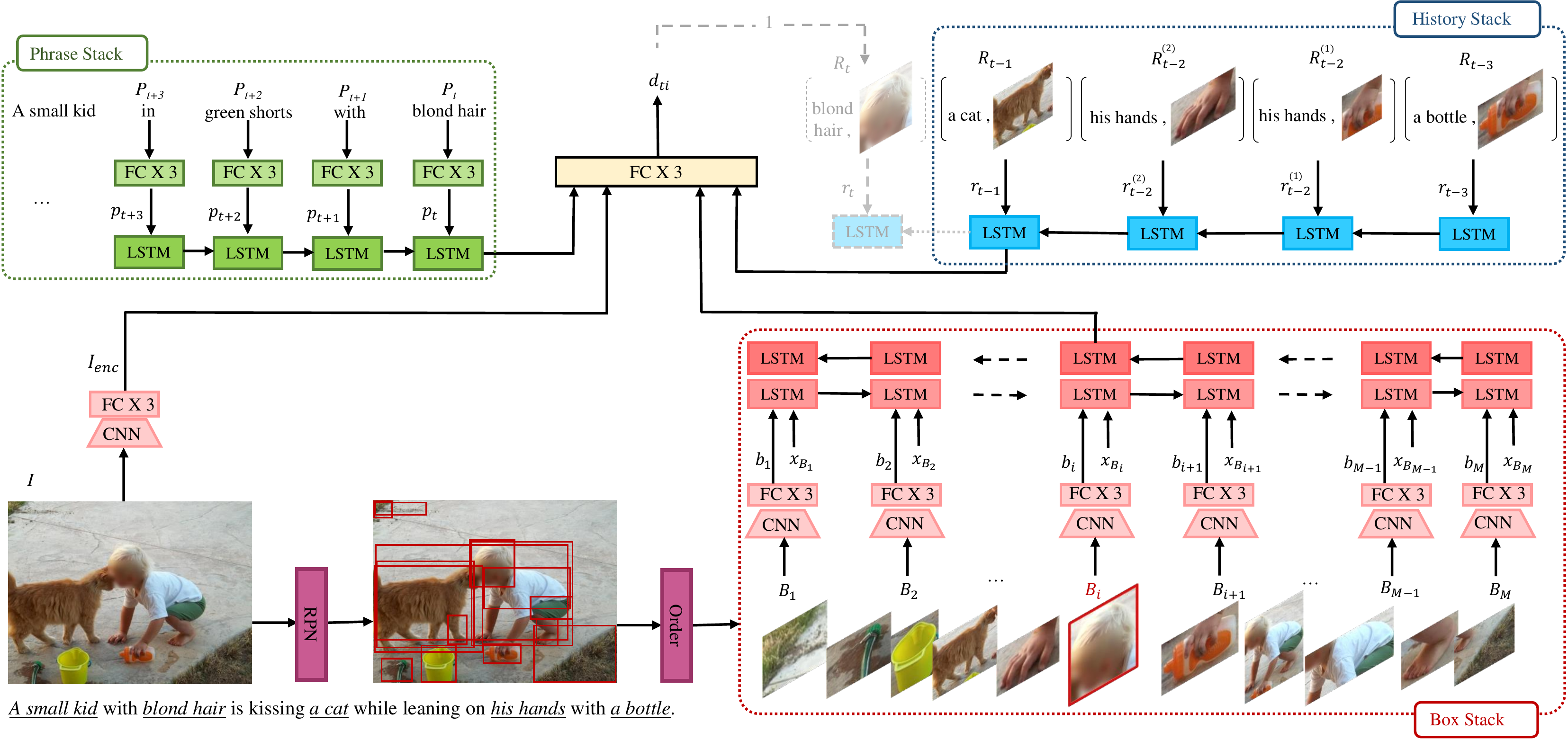}
	\vspace{-5pt}
	\caption{{\bf SeqGROUND neural architecture}. The \textit{phrase stack} contains the sequence of all phrases, not only the noun phrases, yet to be processed in an order and encodes the linguistic dependencies. The \textit{box stack} contains the sequence of bounding boxes that are ordered with respect to their locations in the image. The \textit{history stack} contains the phrase-box pairs that are previously grounded. 
		The grounding decisions for the input phrases are performed sequentially taking into account of the current states of these LSTM stacks in addition to full image representation. The new grounded phrase-box pairs are added to the top of the \textit{history stack}.}
	\label{fig:networkArchitecture}
	\vspace{-7pt}
\end{figure*}

The SeqGROUND framework copes with these challenges by casting the problem as one of
sequential grounding and explicitly representing the state of the entire decision {\em workspace}, including the partially grounded input phrases and boxes. The representation employs LSTM recurrent networks for region proposals, sentence phrases, and the previously grounded content, in addition to dense layers for the full image representation. Figure \ref{fig:networkArchitecture} shows the architecture of our framework. 

We learn a function that maps the state of workspace $\Psi_t$ to a grounding decision $d_{ti}$ for the bounding box $B_i$ at every time step $t$, which corresponds to a decision for phrase $P_t$. 
The decisions $d_{ti}$ manipulates the content of the LSTM networks, resulting in a new state $\Psi_{t+1}$. 
Executing a complete sequence of decisions produces a complete alignment of the input phrases with
the bounding boxes. 
We note, that our model is an extension and generalization of the NeuMATCH framework \cite{dogan2018neural} introduced by Dogan \etal. Further, there is a clear connection with reinforcement learning and policy gradient
methods~\cite{sutton2011reinforcement}. While an RL-based formulation maybe a reasonable future extension,
here we focus on a fully differentiable supervised learning formulation. 

\subsection{Language and Visual Encoders}
\label{subsec:encoder}

We first create encoders for each phrase and each bounding box produced by a region proposal network (RPN). 

\PAR{Phrase Encoder.} The input caption is parsed into phrases $P_1 \ldots P_N$, each of which contains a word or a sequence of words, using \cite{chen2014fast}. We transform each unique phrase into an embedding vector, by performing mean pooling over GloVe \cite{pennington2014glove} features of all its words. This vector is then transformed with three fully connected layers using the ReLU activation function, resulting in the encoded phrase vector $p_j$ for the $j^{\text{th}}$ phrase $(P_j)$ of the input sentence.

\PAR{Visual Encoder.} For each proposed bounding box, we extract features using the activation of the first fully connected layer in the VGG-16 network \cite{simonyan2014very}, which produces a 4096-dim vector per region. This vector is transformed with three fully connected layers using the ReLU activation function, resulting in the encoded bounding box vector $b_i$ for the $i^{\text{th}}$ bounding box $(B_i)$ of the image. The visual encoder is also used to encode the full image $I$ into $I_{enc}$.
\vspace{-10pt}

\subsection{The Grounding Network}
\label{subsec:grounding}
Having the encoded phrases and boxes in the same embedding space, a naive approach for grounding would be maximizing the collective similarity over the grounded phrase-box pairs.
However, doing so ignores the spatial structures and relations within the elements of the two sequences, and can lead to degraded performance. 
SeqGROUND performs grounding by encoding the input sequences and the decision history with stacks of recurrent networks. 
This implicitly allows the network to take into account all grounded as well as ungrounded proposal regions and phrases as 
context for the current grounding decision. We show in the experimental section that this leads to a significant boost in performance.

\PAR{Recurrent Stacks.} 
Considering the input phrases as a temporal sequence, we let the first stack contain the sequence of phrases yet to be processed $P_t, P_{t+1}, \ldots, P_{N}$, at the time step $t$.
The direction of the stack goes from $P_N$ to $P_t$, which allows the information to flow from the future phrases to the current phrase. We refer to this LSTM network as the \textit{phrase stack} and denote its hidden state as $h^P_t$. The input to the LSTM unit is the phrase features in the latent space obtained by the phrase encoder (see Sec.~\ref{subsec:encoder}).

The second stack is a bi-directional LSTM recurrent network that contains the sequence of bounding boxes $B_1, \ldots, B_{M}$ obtained by the RPN. The boxes are ordered from left to right considering their center on the horizontal axis for the forward network\footnote{We experimented with alternative orderings, \eg., max flow computed over pair-wise proposal IoU scores, but saw no appreciable difference in performance. Therefore for cleaner exposition we focus on simpler left-to-right ordering and corresponding results.}. We refer to this bi-LSTM network as the \textit{box stack} and denote its hidden state for the $i^{th}$ box as $h^B_i$. The input to the LSTM unit is the concatenation of the box features in the latent space and the normalized location features $[b_i, x_{b_i}]$.
Note that the state of the box stack does not change with respect to $t$. We keep all the boxes in the stack, since a box that is already used to ground a phrase can be used again to grounding another phrase later on.

The third stack is the \textit{history stack}, which contains only the phrases and the boxes that are previously grounded, and places the last grounded phrase-box pair at the top of the stack. We denote this sequence as $R_1, \ldots, R_L$. 
The information flows from the past to the present. 
The input to the LSTM unit is the concatenation of the two modalities in the latent space and the location features of the box. 
When a phrase $p_j$ is grounded to multiple (K) boxes $b_{\pi(j)} = {b_{(p_j,1)}, \ldots, b_{(p_j,K)}}$, each grounded phrase-box pair becomes a separate input to the LSTM unit, keeping the spatial order of the boxes. For example, the vector $[p_j, b_{(p_j, 1)}, x_{b_{(p_j, 1)}}]$ will be the first vector to be pushed to the top of the history stack for the phrase $p_j$. The last hidden state of the history stack is $h^R_{t-1}$.

The \textit{phrase stack} and \textit{history stack} both perform encoding using a 2-layer LSTM recurrent network, where the hidden state of the first layer, $h^{(1)}_t$, is fed to the second layer:

\begin{subequations}
	\begin{align}
	h^{(1)}_t, c^{(1)}_t &= \text{LSTM}(x_t, h^{(1)}_{t-1}, c^{(1)}_{t-1}) \\
	h^{(2)}_t, c^{(2)}_t &= \text{LSTM}(h^{(1)}_t, h^{(2)}_{t-1}, c^{(2)}_{t-1})
	\enspace ,
	\end{align}
\end{subequations}
where $c^{(1)}_t$ and $c^{(2)}_t$ are the memory cells for the two layers, respectively; $x_t$ is the input for time step $t$.

\PAR{Image Context.} 
In addition to the recurrent stacks, we also provide the encoded full image $I$ to the network as an additional global context.

\PAR{Grounding Decision Prediction.}
At every time step, the state of the three stacks is $\Psi_t = (P_{t^+}, B_{t}, R_{1^+})$ 
, where we use the shorthand $X_{t^+}$ for the sequence $X_t, X_{t+1}, \ldots$ and similarly for $X_{t^-}$. 
The LSTM hidden states can approximately represent $\Psi_t$.
Thus, the conditional probability of grounding decision $d_{ti}$, which represents the decision for bounding box $B_i$ with the phrase $P_t$ is
\begin{equation}
Pr(d_{ti} | \Psi_t ) = Pr(d_{ti} | h^P_t , h^B_i, h^R_{t-1}, I_{enc} ).
\end{equation}
In other words, at time step $t$, a grounding decision is made simultaneously for each box for the phrase at the top of the \textit{phrase stack}. 
Although it may seem that these decisions are made in parallel independently, the hidden states of the \textit{box stack} encode the relation and 
dependencies between all the boxes. 
The above computation is implemented as a sigmoid operation after three fully connected layers on top of 
the concatenated state $\psi_t = [h^P_t, \{h^B_i\}, h^R_{t-1}, I_{enc}]$. ReLU activation is used between the layers.
Further, each positive grounding decision will augment the \textit{history stack}.

In order to ground the entire phrase sequence with the boxes, we apply the chain rule as follows:
\vspace{-3pt}
\begin{equation}
Pr(D_1, \ldots, D_N | \mathcal{P}, \mathcal{B} ) = \prod_{t=1}^N Pr(D_t | D_{(t-1)^-}, \Psi_t )
\end{equation}
\vspace{-3pt}
\begin{equation}
Pr(D_t | \mathcal{P}, \mathcal{B} ) = \prod_{i=1}^M Pr(d_{ti} | D_{(t-1)^-}, \Psi_t ),
\end{equation}

\noindent
where $D_t$ represents the set of all grounding decisions over all the boxes for the phrase $P_t$.
The probability can be optimized greedily by always choosing the most probable decisions.
The model is trained in a supervised manner. From a ground truth grounding of a box and a phrase sequence, we can easily derive the correct decisions, which are used in training. The training objective is to minimize the overall binary cross-entropy loss caused by the grounding decisions at every time step for each $\langle P_t, B_i \rangle$ with $i=1,\ldots,M$.

\PAR{Pre-training.}
As noted in \cite{dogan2018neural}, learning a coordinated representation (or similarity measure) between visual and text data,
while also optimizing a decision network, is difficult. Thus, we adopt 
a pairwise pre-training step to coordinate the phrase and visual encoders to achieve a good initialization for subsequent end-to-end training. 
Note that this is only done for pre-training; the final model is fully differentiable and is fine-tuned end-to-end. 

For a ground-truth pair $(P_k, B_k)$, we adopt an asymmetric similarity proposed by \cite{vendrov2015order}
\begin{equation}
\label{eq:similarity}
F(p_k,b_k) = - ||\max(0, b_k-p_k)||^2
\enspace .
\end{equation}
This similarity function, $F$, takes the maximum value 0, when $p_k$ is positioned to the upper right of $b_k$ in the vector space. 
When that condition is not satisfied, the similarity decreases. In \cite{vendrov2015order}, this relative spatial position defines an entailment relation where $b_k$ entails $p_k$. Here, the intuition is that the image typically contains more information than being described in the text form, so we may consider the text as entailed by the image.  

We adopt the following ranking loss objective by randomly sampling a contrastive 
box $B^\prime$ and a contrastive phrase $P^\prime$ for every ground truth pair. Minimizing the loss function maintains that the similarity of the contrastive pair is below the true pair's by at least the margin $\alpha$:
\vspace{-4pt}
\begin{dmath}
	\label{eq:pairwiseLoss}
	\mathcal{L} = \sum_{i}\left(\mathbb{E}_{b^\prime \neq b_k} \max \left\{0, \alpha-F(b_k, p_k)+F(b^\prime, p_k)\right\}  \\
	+ \mathbb{E}_{p^\prime \neq p_k}  \max\left\{0, \alpha-F(b_k, p_k)+F(b_k, p^\prime)\right\} \right)
\end{dmath}
Note the expectations are approximated by sampling.

\section{Experiments}
\label{sec:experiments}
\subsection{Setup and Training.}
\begin{figure*}[!htb]
	\vspace{-20pt}
	\centering 
	\begin{tabular}[b]{c}
		\subfloat[][]{\adjustbox{raise=-5.8pc}{\includegraphics[height=4.9cm]{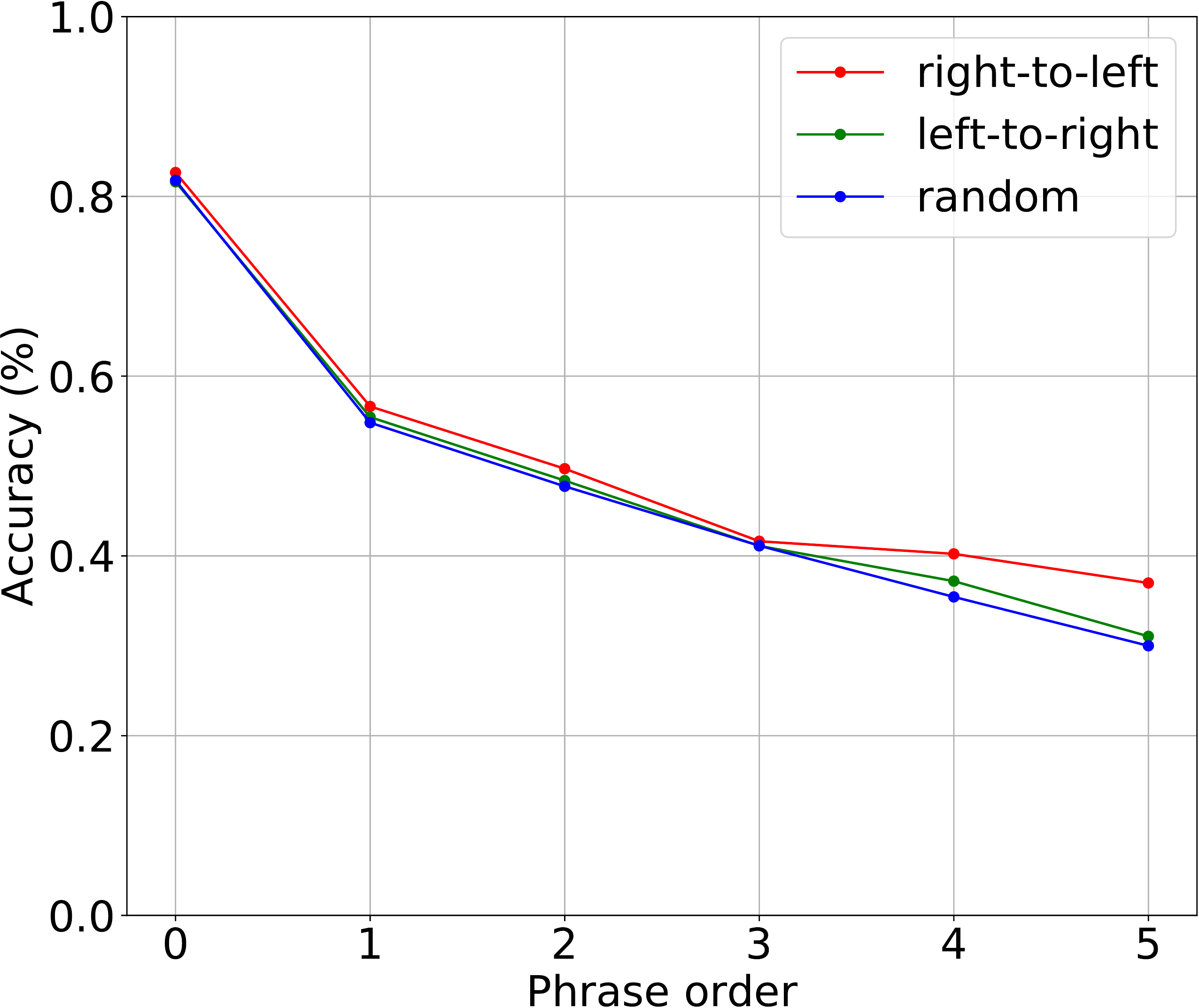} \label{fig:plot}}}
		\subfloat[][]{
			\resizebox{0.55\textwidth}{!}{
				\begin{tabular}{lccccr}
					\toprule
					& \multicolumn{4}{c}{\textbf{Components}} & \\ 			\cmidrule(lr){2-5}
					& Visual context& Bounding box & Phrase & History & \textbf{Accuracy} \\ \midrule
					MSB & none & simple & simple & none & 43.85 \\
					MSBs & none & simple & simple & none & 50.90 \\ \midrule
					NH & global & bi-LSTM & LSTM & none & 59.55\\
					NI & none & bi-LSTM & LSTM & LSTM & 60.34 \\
					SPv & global & bi-LSTM & simple & LSTM & 57.94 \\
					SBv & global & simple & LSTM & LSTM & 55.68 \\
					SPvBv & global & simple & simple & LSTM & 53.75\\
					SBvPvNH & global & simple & simple & none & 52.91 \\ \midrule
					SeqGROUND & global & bi-LSTM & LSTM & LSTM & \textbf{61.60} \\ 
					\bottomrule
				\label{tab:ablations}
				\end{tabular}}}
			\end{tabular}
			\vspace{-9pt}
			\caption{{\bf The performance of various design choices.} \protect\subref{fig:plot} Grounding accuracy versus the ordering of the grounded phrase among the noun phrases of the sentence. Red, green, and blue plots show the performance when the phrases to the  LSTM cell are ordered left-to-right (lexical order), right-to-left (reverse lexical order), and randomly, respectively. \protect\subref{tab:ablations} Grounding accuracy of baselines and ablated models. }
			\vspace{-8pt}
		\end{figure*}

We use Faster R-CNN~\cite{ren2015faster} as an underlying bounding box proposal mechanism with ResNet50 as the backbone. The extracted bounding boxes are then sorted from left-to-right by their central x-coordinate to be fed into the Bi-LSTM network of the \textit{box stack}. This way, the objects appearing close tend to be represented closer together, so that the \textit{box stack} can represent the overall context better. 
Following the prior works (see Tab.~\ref{tab:flickr_results}), we assume that the noun phrases that are to be grounded have already been extracted from the descriptive sentences. We also use the intermediate words of the sentences together with the given noun phrases in the \text{phrase stack} to preserve the linguistic structure; this also results in a more complex train/test scenario. 

SeqGROUND is trained in two stages that differ in \textit{box stack} input. In the first stage, we only feed the groundtruth instances to the \textit{box stack}, which are coming from the dataset annotation, for an image. The boxes that have the same label as the phrase are considered as positive samples, while the remaining boxes as negative samples. This set-up provides an easier phrase grounding task due to the low number of input boxes which are contextually distinct and well-defined without being redundant. Thus, it provides a good initialization for the second stage where we use the box proposals by the RPN. 

For the second stage, we map each bounding box, coming from the RPN, to the groundtruth instances with which it has IoU overlap equal to or greater than 0.7, and label them as positive samples for the current phrase. 
The remaining proposed boxes having IoU overlap less than 0.3 with the groundtruth instances are labeled as negative samples for that phrase. The labeled positive and negative samples are sorted and then fed into the Bi-LSTM network. 
It is possible to optimize for the loss function of all labeled boxes, but this will bias towards negative samples as they dominate. Instead, we randomly sample negative samples that contribute to the loss function in a batch, where the sampled positive and negative boxes have a ratio of 1:3. If the number of negative samples within a batch is not enough, we let all the samples in that batch contribute to the loss.
In this way, the spatial context and dependencies are represented without gaps by the Bi-LSTM unit of the \textit{box stack}, while preventing biasing towards negative grounding decisions.
After the second stage of training, we adopt the standard hard negative mining method ~\cite{felzenszwalb2010object, sung1998example} with a single pass on each training sample.

At test time, we use all the proposed boxes to feed them to the \textit{box stack} after ordering them with respect to their locations. When multiple boxes are grounded to the same phrase, we apply non-maximum suppression with an IoU overlap threshold of 0.3, which is tuned on the validation set. In this way, multiple box results for the same instance of a phrase are discarded, while the boxes for different instances of the same phrase are kept.
\vspace{-2pt}
\subsection{Datasets and Metrics} 
We evaluate our approach on the Flickr30K Entities dataset \cite{plummer2015flickr30k} which contains $31,783$ images, each annotated with five sentences. For each sentence, the noun phrases are provided with their corresponding bounding boxes in the image. We use the same training/validation/test split as the prior work, which provides $1,000$ images for validation, $1,000$ for testing, and $29,783$ images for training. It is important to note that a single phrase can have multiple groundtruth boxes, while a single box can match multiple phrases within the same sentence. 
Consistent with the prior work, we evaluate SeqGROUND with the ground truth bounding boxes. If multiple boxes are associated with a phrase, we represent the phrase as the union of all its boxes on the image plane. Following the prior work, successful grounding of a phrase requires predicted area to have at least 0.5 IoU (intersection over union) with the groundtruth area. Based on this criteria, our measure of performance is grounding {\em accuracy}, which is the ratio of correctly grounded noun phrases..  

\subsection{Baselines and Ablation Studies}
\label{subsec:ablations}
In order to understand the benefits of the individual components of our model, we perform an ablation study where certain stacks are either removed or modified. 
The model \textit{NH} lacks the \textit{history stack} where the previously grounded phrase-box pairs do not affect the decisions for the upcoming phrases in a sentence. 
The model \textit{NI} lacks the full image context where the only visual information to the framework is the \textit{box stack}. 
The model \textit{SBv} (simple box vector) lacks the bi-LSTM network for the boxes, and direclty uses the encoded box features coming from the triple fully connected layers in Figure~\ref{fig:networkArchitecture}. In this way, the decision for a phrase-box pair is made independently of the other box candidates. 
The model \textit{SPv} (simple phrase vector) lacks the LSTM network for the \textit{phrase stack} and directly uses the encoded phrase features coming from the triple fully connected layers in Figure~\ref{fig:networkArchitecture}. In this design, the framework is not aware of the upcoming phrases so that the decision for a phrase-box pair is made without the linguistic relations.
Similarly, \textit{SPvBv} lacks the bi-LSTM and LSTM networks for the box and phrase stacks, respectively. Moreover, \textit{SPvBvNH} lacks the history module as an addition. 

\begin{table}
	\centering
	\begin{tabular}{lrrr}
		\toprule
		Method & & & Accuracy \\ \midrule
		SMPL~\cite{wang2016structured} & & & 42.08 \\
		NonlinearSP~\cite{wang2016learning} & & & 43.89 \\
		GroundeR~\cite{rohrbach2016grounding} & & & 47.81 \\
		MCB~\cite{fukui2016multimodal} & & & 48.69 \\
		RtP~\cite{plummer2015flickr30k} & & & 50.89 \\
		Similarity Network~\cite{wang2018learning} & & & 51.05 \\
		RPN+QRN~\cite{chen2017query} & & & 53.48 \\
		IGOP~\cite{yeh2017interpretable} & & & 53.97 \\
		SPC+PPC~\cite{plummer2017phrase} & & &  55.49 \\ 
		SS+QRN~\cite{chen2017query} & & & 55.99 \\
		CITE~\cite{plummer2018conditional} & & &  59.27 \\ \midrule
		SeqGROUND & & & \textbf{61.60} \\ 
		\bottomrule
	\end{tabular}
 	\vspace{-4pt}
	\caption{{\bf Phrase grounding accuracy} (in percentage) of the state-of-the-art methods on the Flickr30k Entities dataset.}
	\label{tab:flickr_results}
 	\vspace{-8pt}
\end{table}

Moreover, we created a baseline that performs phrase grounding in a non-sequential way by picking the most similar bounding box in the joint embedding space. To encode the phrases and boxes, we used the same phrase-visual encoders that were pre-trained in Section~\ref{subsec:grounding}. For each image-sentence input, we created a similarity matrix for all possible phrase-box pairs using the similarity function \ref{eq:similarity}. Using this matrix, the phrases were grounded to the most similar box and boxes for the models \textit{MSB} and \textit{MSBs}, respectively. 

Table~\ref{tab:ablations} shows the performance of the six ablated models and two baselines on the Flickr30K Entities dataset. 
All these models perform substantially worse than the complete model of SeqGROUND. 
This confirms our intuition that knowing the global context for both visual and textual data, in addition to history and future, plays an important role in phrase grounding. We conclude that each stack contributes to our full model's superior performance. 

\PAR{Phrase Ordering.}
We consider several ways of ordering the phrases of a sentence. 
\begin{enumerate}[topsep=0pt,itemsep=0pt,partopsep=0pt, parsep=0pt]
	\item{Left-to-Right: The network grounds the phrases in lexical order, starting from the first phrase of the sentence.}
	\item{Right-to-Left: The network grounds the phrases in reverse lexical order, starting from the last phrase.}
	\item{Random: We randomly order the phrases for the \textit{phrase stack}, and keep the ordering fixed for all of the training.} 
\end{enumerate}
At test time, the phrases are ordered in the same order as the corresponding design's training time. 
The grounding accuracy with respect to the phrase's order among the noun phrases of the sentence is shown in Figure~\ref{fig:plot} for different ordering options. For all ordering options, the accuracy for the first phrase is significantly higher than the others. 
This is due to the fact that the first phrases usually belong to the category of \textit{people} or \textit{animals} which have significantly more samples in the dataset. Moreover, the candidate boxes from RPN are more accurate in proposing boxes for these categories which provides easier detection. The grounding accuracy drops towards the last phrases, which usually belong to the categories that have less samples in the dataset. Ordering the phrases  right-to-left boosts the performance slightly for the last phrases of the sentence, since they are the first ones to be grounded. 
In this way, these hard-to-ground phrases are not a subject of a possible error cumulation in the \textit{history stack}. 

\PAR{Unguided Testing.} SeqGROUND does not necessarily need to be given phrases to ground. Due to its sequential nature, it scans through all the phrases in the sentences, selected phrases or not, and makes decisions which of those to ground and where (see Fig.~\ref{fig:image_results}). The network implicitly learns to distinguish entities to-be-grounded during training. This is a more complex scenario than addressed by prior works, which only focus on phrases that implicitly have groundings. The results in Table~\ref{tab:flickr_results}, \ref{tab:category_results}, and Figure~\ref{fig:image_results} are obtained via unguided testing, which is a key property of our method.

\subsection{Results}

We report the performance of SeqGROUND on the Flickr30K Entities dataset, and compare it with the state-of-the-art methods\footnote{Performance on this task can be further improved by using Flickr30K-tuned features to represent the image regions, with the best result of 61.89\% achieved by CITE  \cite{plummer2018conditional}. Futhermore, the use of an integrated proposal generation network to learn regression over Flickr30K Entities improves the result up to 65.14\% as achieved by \cite{chen2017query}.} in Table~\ref{tab:flickr_results}. 
SeqGROUND is the top ranked method in the list, improving the overall grounding accuracy by 2.33\% to 12.91\% by performing phrase grounding as a sequential and contextual process, compared to the prior work.
For a fair comparison, all these methods use a fixed RPN to obtain the candidate boxes and represent them in features that are not tuned on the Flickr30K Entities dataset.  
We believe that using an additional conditional embedding unit as in~\cite{plummer2018conditional}, and the integration of a proposal generation network with a spatial regression that is tuned on Flickr30K Entities as in~\cite{chen2017query} should improve the overall result even more. 
Table~\ref{tab:category_results} shows the phrase grounding performance with respect to the coarse categories in Flickr30K Entitites dataset. Competing results are directly taken from the respective papers, if applicable.

\begin{table*}
	\centering
	\resizebox{0.8\textwidth}{!}{
		\begin{tabular}{lcccccccc}
			\toprule
			Method & people & clothing & body parts & animals & vehicles & instruments & scene & other \\ \midrule
			SMPL~\cite{wang2016structured} & 57.89 & 34.61 & 15.87 & 55.98 & 52.25 & 23.46 & 34.22 & 26.23 \\
			GroundeR~\cite{rohrbach2016grounding} & 61.00 & 38.12 & 10.33 & 62.55 & 68.75 & 36.42 & 58.18 & 29.08 \\
			RtP~\cite{plummer2015flickr30k} & 64.73 & 46.88 & 17.21 & 65.83 & 68.72 & 37.65 & 51.39 & 31.77 \\
			IGOP~\cite{yeh2017interpretable} & 68.71 & 56.83 & 19.50 & 70.07 & 73.72 & 39.50 & 60.38 & 32.45 \\
			SPC+PPC~\cite{plummer2017phrase} & 71.69 & 50.95 & 25.24 & 76.23 & 66.50 & 35.80 & 51.51 & 35.98 \\
			CITE~\cite{plummer2018conditional} & 73.20 & 52.34 & 30.59 & 76.25 & 75.75 & 48.15 & 55.64 & 42.83 \\ \midrule
			SeqGROUND & 76.02 & 56.94 & 26.18 & 75.56 & 66.00 & 39.36 & 68.69 & 40.60 \\ 
			\bottomrule
		\end{tabular}
	}
	\vspace{-5pt}
	\caption{{\bf Comparison of phrase grounding accuracy} (in percentage) over coarse categories on Flickr30K dataset.}
	\label{tab:category_results}
	\vspace{-7pt}
\end{table*}
\begin{figure*}
	\centering
	\includegraphics[width=1.8\columnwidth]{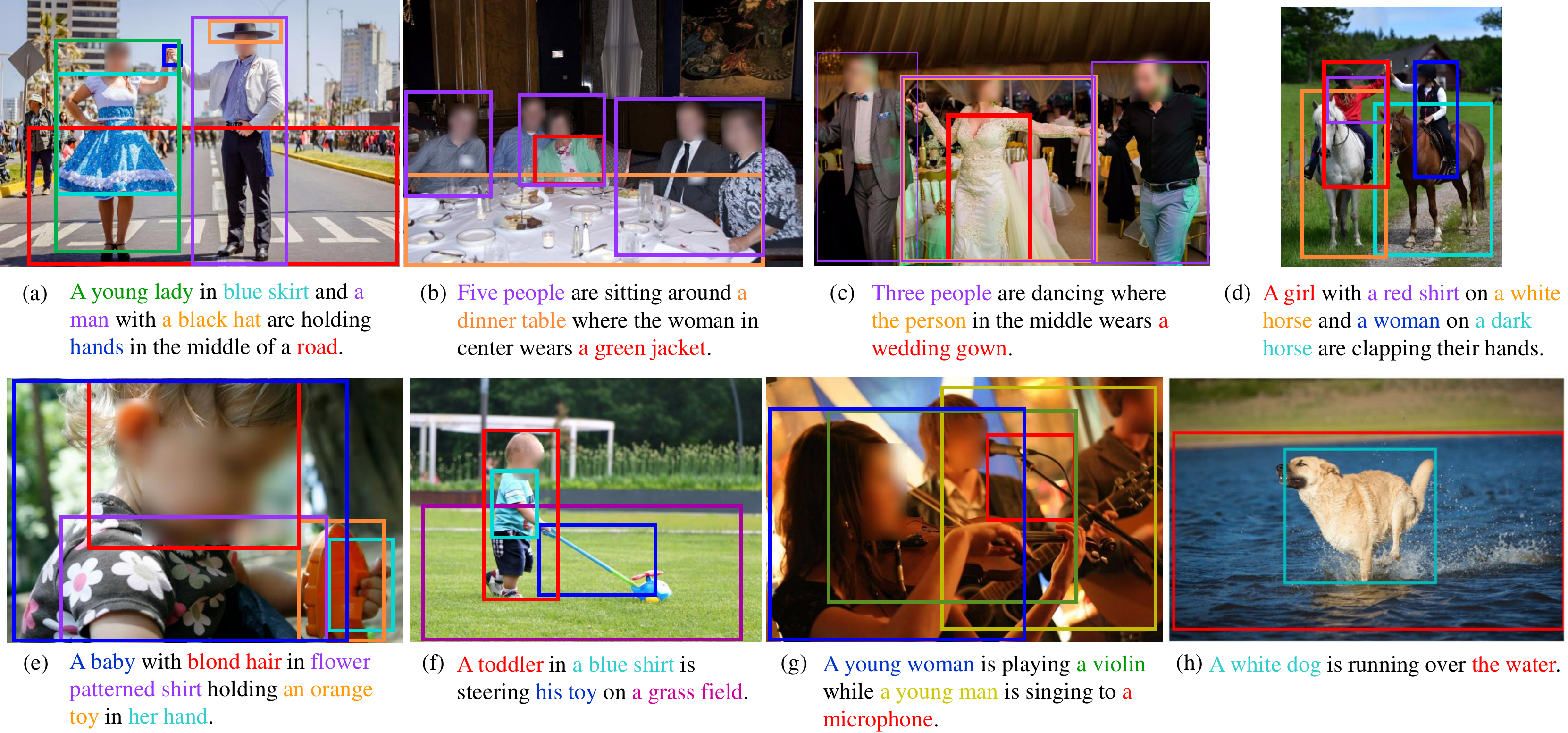}
	\vspace{-6pt}
	\caption{{\bf Sample phrase grounding results obtained by SeqGROUND}\protect\footnotemark. The colored bounding boxes show the predicted grounding of the phrases in the same color. See text for discussion.}
	\label{fig:image_results}
	\vspace{-7pt}
\end{figure*}

We show some qualitative results in Figure~\ref{fig:image_results} to highlight the capabilities of our method in challenging scenarios. 
In (a) and (e), we see a successful grounding of long sequence of phrases, note the correct grounding of \textit{hands} in (a) despite other \textit{hands} candidates. 
In (b), phrases are correctly grounded to multiple boxes, instead of one large single box for \textit{five people} which would contain mostly the \textit{dinner table}.
Likewise, (c) shows an example where a single box is used to ground multiple phrases, \textit{three people} and \textit{the person} which are positioned far apart. Phrase grounding with many-to-many matching is one of the distinguishing properties of SeqGROUND, which is partially or completely missing in most of the competing methods.
In (d), SeqGROUND could distinguish which boxes to ground the phrases \textit{a girl} and \textit{a woman}, suppressing the other candidates despite their similar context. We believe this is possibly due to SeqGROUND's ability to perform in a sequential way where it consders the global image and text context. As an intuitive example, the performed grounding starts by matching \textit{a dark horse} to the correct box. Encoding this grounded pair and the overall contextual information, it grounds \textit{a woman} to the correct box, which is just above \textit{a dark horse}, instead of getting confused by the box that has \textit{A girl}. At the decision time for \textit{a woman}, the \textit{phrase stack} encodes the future information, which is \textit{a girl} should have a \textit{red shirt} and should be on \textit{a white horse}. Taking account of this information likely has led SeqGROUND to eliminate the box for \textit{a girl} at the decision time for \textit{a woman}. 

All these images, and more in the supplementary material, show state-of-the-art performance of SeqGROUND due to its contextual and sequential nature.

\section{Conclusions}
In this paper, we proposed an end-to-end trainable Sequential Grounding Network (SeqGROUND) that formulates grounding of multiple phrases as a sequential and contextual process. SeqGROUND encodes region proposals, and all phrases into two stacks of LSTM cells along with the partially grounded phrase-region pairs to perform the grounding decision for the next phrase. Results on the Flickr30K Entities benchmark dataset and ablations studies show significant improvements of this model over more traditional grounding approaches.

\footnotetext{Due to copyright issues of the images in the Flickr30K Entities dataset, we are not allowed to show images from it. Instead, we created similar content with images that have Creative Commons license, and blurred the faces of people due to privacy concerns.}
\clearpage
\balance
{\small
\bibliographystyle{ieee}
\bibliography{egbib}
}

\clearpage

\onecolumn
\centerline{\Large \textbf{Supplementary Material}}
\vspace{1.5cm}

In this supplementary material, we provide additional details on implementation of our framework, such as parameters, dimensions, etc. Furthermore, we will provide more phrase grounding results computed by our approach that require many-to-many matching of phrases and bounding boxes.
\vspace{10pt}
\PAR{Additional Implementation Details} \\
\\
The phrase and the visual encoder represent the input phrases and regions (boxes) in a $500$-dimensional joint embedding space. 
The phrase encoder, which is shown by \textit{3xFC} in green in Figure 2 of the main paper, has one dropout layer with a rate of $0.4$ that is followed by three fully connected layers with ReLu activations, and a normalization layer. These fully connected layers have output dimensionality of $1500$, $1000$, and $500$ respectively. The visual encoder, which is shown by \textit{3xFC} in pink in Figure 2 of the main paper has the same layout as well. The batch size is set to 32, which provides 31 contrastive samples for every positive pair. The pre-training step is performed with the Adam optimizer using a learning rate of $10^{-4}$, and a gradient clipping threshold of $2.0$.

The \textit{phrase stack}and the \textit{history stack} are implemented with a two-layer LSTM network with an output space dimensionality of $500$. Each layer of the bidirectional LSTM network of the \textit{box stack} has the same output dimensionality, too. Between each layer in the stacks, there is a dropout layer with a rate of $0.25$. Another dropput layer with the rate of $0.4$ is placed after concatenating the hidden states of the stacks, which is before the fully connected layers.  The last fully connected layer that outputs the decision prediction has a sigmoid activation, in order to train the model with binary cross-entropy loss. The batch size is set to 10 image-sentence pairs. The full model is trained with the Adam optimizer using a learning rate of $10^{-3}$ and $10^{-4}$, respectively for the two stages of training mentioned in Section $4.1$ of the  main paper. At test time, our method performs in 1.39 seconds, on average, for an image-sentence pair. Training (including pre-training and the stages in Sec. 4.1) is close to 4 days, using one NVIDIA Titan X.

\end{document}